\documentclass[runningheads]{llncs}
\usepackage[T1]{fontenc}
\usepackage{graphicx,verbatim}
\usepackage{amsmath}
\usepackage{amssymb}

\usepackage{booktabs}

\usepackage{url}
\usepackage{hyperref}
\usepackage{color}

\newcommand{\pmnum}[2]{#1\tiny{$\pm$#2}}

\usepackage{fancyhdr}
\begin{document}
\title{Anisotropic Fourier Features for Positional Encoding in Medical Imaging}
%

\author{
Nabil Jabareen\inst{1}
\and
Dongsheng Yuan\inst{1}\inst{2}
\and
Dingming Liu\inst{1}
\and
Foo-Wei Ten\inst{1}
\and
S\"oren Lukassen \inst{1}
}

%
\institute{Center of Digital Health, Berlin Institute of Health at Charit\'e – Universit\"atsmedizin Berlin, Germany\\
\email{\{nabil.jabareen,soeren.lukassen\}@charite.de}\and
Department of Experimental Neurology, Charit\'e – Universit\"atsmedizin Berlin, corporate member of Freie Universität Berlin and Humboldt-Universit\"at
zu Berlin, Germany}

\maketitle              
\begin{abstract}
The adoption of Transformer-based architectures in the medical domain is growing rapidly.
In medical imaging, the analysis of complex shapes - such as organs, tissues, or other anatomical structures - combined with the often anisotropic nature of high-dimensional images complicates these adaptations.
In this study, we critically examine the role of Positional Encodings (PEs), arguing that commonly used approaches may be suboptimal for the specific challenges of medical imaging.
Sinusoidal Positional Encodings (SPEs) have proven effective in vision tasks \cite{dosovitskiy_image_2021}, but they struggle to preserve Euclidean distances in higher-dimensional spaces \cite{li_learnable_2021}.
Isotropic Fourier Feature Positional Encodings (IFPEs) have been proposed to better preserve Euclidean distances \cite{li_learnable_2021}, but they lack the ability to account for anisotropy in images.
To address these limitations, we propose Anisotropic Fourier Feature Positional Encoding (AFPE), a generalization of IFPE that incorporates anisotropic, class-specific, and domain-specific spatial dependencies.
We systematically benchmark AFPE against commonly used PEs on multi-label classification in chest X-rays, organ classification in CT images, and ejection fraction regression in echocardiography.
Our results demonstrate that choosing the correct PE can significantly improve model performance.
We show that the optimal PE depends on the shape of the structure of interest and the anisotropy of the data.
Finally, our proposed AFPE significantly outperforms state-of-the-art PEs in all tested anisotropic settings.
We conclude that, in anisotropic medical images and videos, it is of paramount importance to choose an anisotropic PE that fits the data and the shape of interest.

All of our code is open source and publicly available\footnote{\url{https://github.com/NabJa/AFPE}}.

\keywords{Positional Encoding \and Anisotropy \and  Fourier Features \and Vision Transformer}

\end{abstract}

\section{Introduction}

Transformer architectures have revolutionized computer vision, consistently outperforming traditional Convolutional Neural Networks (CNNs) across a wide range of medical imaging tasks \cite{hatamizadeh_unetr_2022,willemink_toward_2022,chan_super-resolution_2023,khader_multimodal_2023,smith_vision_2024}.
Unlike CNNs, which inherently encode spatial relationships through local receptive fields and weight sharing, Vision Transformers (ViTs) rely on Positional Encodings (PEs) to incorporate spatial context \cite{dosovitskiy_image_2021}.
This fundamental architectural difference raises important questions about the optimal design of PEs for medical imaging applications, where spatial and geometric properties critically influence diagnostic interpretation.

Unlike natural images or language sequences, medical imaging modalities often capture complex 3D anatomical shapes and dynamic physiological processes with intrinsic spatial anisotropy.
For example, volumetric MRI or CT scans exhibit differing in-plane and through-plane resolutions, while echocardiography videos encode both spatial and temporal dimensions with distinct characteristics \cite{wang_chestx-ray8_2017,yang_medmnist_2023,ouyang_video-based_2020}.
Moreover, anatomical structures exhibit conserved spatial relationships and shape continuity that are biologically meaningful and stable across populations \cite{mensah_establishing_2015,sakai_thoracic_2020}.
These domain-specific spatial and geometric constraints motivate the development of PEs that respect shape and anisotropy, enabling Transformers to better model anatomical variability and pathology.

\begin{figure}[h]
    \centering
    \includegraphics[width=\textwidth]{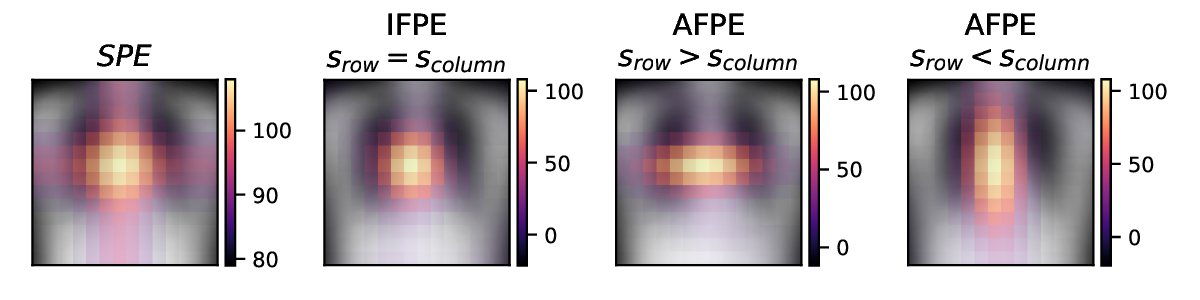}
    \caption{Anisotropic Fourier Feature Positional Encoding (AFPE) captures directional spatial dependencies more effectively than Sinusoidal Positional Encoding (SPE) and Isotropic Fourier Feature Positional Encoding (IFPE). Shown are dot product similarity maps on a Vision Transformer-style patch grid, with similarities computed relative to the central patch. Note that SPE does not capture Euclidean distances along diagonal directions between patches. The background image shows the mean of all ChestX training images, highlighting anatomical consistency across the dataset.}
    \label{fig:iso-pe-comparison-2d}
\end{figure}

Traditional Sinusoidal Positional Encodings (SPEs) have been widely used in ViTs but suffer from limitations in preserving Euclidean distances beyond one-dimensional sequences \cite{li_learnable_2021}.
Isotropic Fourier Feature Positional Encodings (IFPEs), which embed coordinates into higher-dimensional spaces using random Fourier bases, have been shown to effectively capture spatial similarity in general vision tasks \cite{tancik_fourier_2020} and are a key component of prominent architectures such as the Segment Anything Model (SAM) \cite{kirillov_segment_2023}. 
Anatomical shapes and their functional dynamics exhibit directional continuity and anisotropic spatial dependencies that are not fully captured by isotropic PEs.
This gap motivates our proposal of Anisotropic Fourier Feature Positional Encoding (AFPE), which generalizes IFPE by introducing dimension-specific scaling factors to explicitly model anisotropy aligned with physical pixel or voxel spacing and anatomical shapes.
Figure~\ref{fig:iso-pe-comparison-2d} illustrates the improved spatial similarity properties of IFPE over SPE, as well as AFPEs' ability to capture directional continuity.

AFPE is designed to focus on the integration of shape-aware and geometry-informed learning in medical imaging.
In this paper, we demonstrate that leveraging AFPE enhances the representation of spatial structure and shape continuity in ViT models, leading to improved classification and regression performance in anisotropic settings.

\section{Positional Encodings}
In the vanilla Transformer architecture, token sequences are processed using a permutation-invariant attention mechanism \cite{vaswani_attention_2017}.
Due to this permutation invariance, PEs are added to each token to encode its position within the sequence, enabling the model to capture the order-dependent structure of the data.
The ViT applies the same logic to computer vision tasks, by using flattened and linearly projected image patches as tokens \cite{dosovitskiy_image_2021}.
A PE is then added by element-wise addition to these tokens.
The PE is generated via a function \(f:p\rightarrow \mathbb{R}^D\) that maps a \(m\)-dimensional position \(p\) to a vector \(d=(d_1,d_2,\dots,d_D) \in R^D\), where \(D\) is the dimensionality of the generated vector and matches the dimensionality of the tokens.

\subsection{Sinosoidal Positional Encoding (SPE).}
In the original Transformer \cite{vaswani_attention_2017}, the non-learnable SPE was proposed for the one-dimensional position $p$:
\begin{equation}
    \label{eq:spe}
    SPE_t(p) = \begin{cases}
        \sin(\frac{p}{t^{freq_j}}) & \text{if } j \text{ mod } 2 = 0 \\
        \cos(\frac{p}{t^{freq_j}}) & \text{else}
    \end{cases}
\end{equation}
where \(d_j\) corresponds to the \(j\)-th index within the encoding vector \(d\) with \(j=1,2,\dots,D\).
The temperature $t$ and the base frequency \(freq_j = \frac{2d_j}{D}\) correspond to different constant frequencies of the sinusoidal functions.
In the ViT the SPE is applied to higher spatial dimensions, like images or videos, by encoding every spatial or temporal dimension separately and then concatenating them \cite{dosovitskiy_image_2021}.

\subsection{Isotropic Fourier Feature Positional Encoding (IFPE).}
In the seminal work of Rahimi and Rech \cite{rahimi_random_2007}, Fourier feature mappings have been first proposed to approximate a stationary kernel function. 
Building on this work, Tancik et al. have shown that using Fourier features let networks learn high frequency functions in low dimensional domains like encoding positions \cite{tancik_fourier_2020}.
This leads to the formulation of an Isotropic Fourier Feature Positional Encoding (IFPE) for a \(m\)-dimensional position \(p\), that can be generated using:
\begin{equation}
    \label{eq:fpe}
    IFPE_s(p) = [\sin(2\pi B_s p) || \cos(2\pi B_s p)]^T ,
\end{equation}
where \(B_s \in \mathbb{R}^{m \times D/2}\) is sampled from a Gaussian distribution $B_s \sim \mathcal{N}(0,s)$ \cite{tancik_fourier_2020}.
The $||$ refers to a concatenation along the feature dimension $D$.
The variance s is a hyperparameter that controls the spread (or concavity) of the similarity distribution \cite{tancik_fourier_2020}.
Specifically, a larger value of s results in more distant positions being more similar to each other.

\subsection{Learnable Fourier Feature Positional Encoding (LFPE).}
The IFPE uses a Gaussian distribution \(B\) for a Fourier feature mapping.
However, this distribution does not have to be strictly Gaussian and can be learned as additional model parameters \cite{li_learnable_2021}.
This involves allowing gradients to flow through the distribution during training.
Analogous to the IFPE, we define the LFPE as follows:
\begin{equation}
    \label{eq:lfpe}
    LFPE(p)=[\sin(2 \pi Wp)  || \cos(2\pi Wp)]^T,
\end{equation}
where \(W \in \mathbb{R}^{m \times D/2}\) are trainable parameters, initialized with a standard Gaussian distribution \(W \sim \mathcal{N}(0, 1)\).

\subsection{Anisotropic Fourier Feature Positional Encoding (AFPE).}
    It has been shown, that the IFPE scale hyperparameter \(s\) is tunable and can influence model performance \cite{tancik_fourier_2020,li_learnable_2021}.
    However, since the scale affects all spatial dimensions equally, the anisotropic nature of images or videos can not be captured.
    We propose to generalize IFPE, by treating the scale \(s\) for every spatial or temporal dimension separately.
    This can be achieved by sampling a different Gaussian distribution \(B_i\) for every spatial dimension \(m\) in a position \(p\):
    \begin{equation}
    \label{eq:afpe}
    B_{s_i} \sim \mathcal{N}\left(0, s_i\right) \quad \text{for all } i \in {1, \dots, m},    \end{equation}
    where \( s = (s_1, s_2, \dots, s_m)^\top \in \mathbb{R}^m \) is a vector where each \( s_i > 0 \) represents the scale for the \( i \)-th spatial dimension.
    Finally, this Gaussian distribution is used as in the formulation of the IFPE (Eq. \ref{eq:fpe}).
The effect of various scales on the resulting AFPE can be seen in Figure~\ref{fig:iso-pe-comparison-2d}.

\subsection{Additional Baselines.}
\label{sec:baselines}
Instead of adding a spatial inductive prior by handcrafted PEs, the model can learn the PE from scratch.
One method for learning PEs is to randomly initialize them and update them concurrently with all other model parameters through gradient descent.
This approach will be referred to as \textbf{\textit{Learnable}}.
As additional simple baseline, all models were trained without any PE and will be referred to as \textbf{\textit{None}}.

\section{Experimental Setup}
To evaluate the effectiveness of different PEs, we conducted experiments on five diverse medical imaging datasets, spanning multi-label and multi-class classification as well as regression tasks, under varying degrees of anisotropy.
Anisotropy was defined as the ratio between the spatial resolutions of the input image dimensions.
Our experiments were designed to assess the impact of incorporating structural priors and anisotropy-aware PEs on model performance across diverse tasks, imaging modalities, and target labels.

First, we investigated whether a structural prior related to disease shape can be integrated into a classification pipeline using the ChestX dataset.
To this end, the scale parameter of AFPE was optimized per class, and the class-specific shape was compared to the corresponding AFPE representation.

Second, we examined whether AFPE improves image classification in settings where anisotropy is a significant factor.
This was tested on OrganMNIST3D under different levels of increasing anisotropy.

Third, we explored the performance of AFPE on an ejection fraction (EF) regression task using EchoNet.
This complex spatio-temporal task requires the model to identify relevant frames for estimating end-diastolic and end-systolic volumes (EDV and ESV), which are then used to compute the EF.
Here, the goal was to determine whether PEs can retain their effectiveness when space and time are inherently independent dimensions.

Finally, we compared the shape-descriptive capabilities of SPE, IFPE, and AFPE using the AdrenalMNIST3D and VesselMNIST3D datasets.
This was achieved by predicting Feret’s minimum and maximum diameter of anatomical structures under varying anisotropy conditions.

\subsection{Datasets}
\label{sec:dataset-tasks}

\subsubsection{ChestX.}
The NIH Chest X-ray dataset\footnote{\url{https://nihcc.app.box.com/v/ChestXray-NIHCC}} includes two-dimensional X-ray images, with a training set of 78,506 images, a validation set of 12,533, and a test set of 21,081 \cite{wang_chestx-ray8_2017}.
The dataset consists of 43.5\% female and 56.5\% male patients, with an average age of 46.6$\pm$16.6 years (mean$\pm$standard deviation).
The images were downsampled using bilinear interpolation from \(1024^2\) pixels to \(224^2\) pixels.
We used a patch size of \(16^2\), resulting in a grid of \(14^2\) image patches.
This task is a multi-label classification task to predict 19 different diseases and healthy status.
Due to the large class imbalance in this dataset we chose the area under the precision-recall curve (AUPRC) as a validation metric \cite{maier-hein_metrics_2024}.
The ViTs trained on this dataset were trained with 12 layers, 12 heads and a Multi-Layer Perceptron (MLP) dimension of 768 resulting in 42.7 million trainable parameters.

\subsubsection{MedMNIST3D.} We utilized three datasets from the MedMNIST3D collection\footnote{\url{https://medmnist.com/}}.
These include CT scans capturing various body organs (OrganMNIST3D), shape masks of adrenal glands derived from abdominal CTs (AdrenalMNIST3D), and three-dimensional vascular models of the brain obtained through mesh reconstructions from magnetic resonance angiography images (VesselMNIST3D) \cite{yang_medmnist_2023}.
All images have a shape of $64^3$ voxels and a $1mm^3$ voxel spacing.
Anisotropy was simulated in the axial plane by omitting voxels at regular spatial intervals.
Due to the limited dataset size, we reduced the ViT model to only 2 layers, 4 attention heads, and an MLP dimension of 288, resulting in 474,907 trainable parameters.
The patch size was set to \(4^3\), yielding \(16^2\) patches in the isotropic case.
The input and patch sizes for different anisotropy levels are listed in Table~\ref{tab:main-table}.

For both shape datasets, VesselMNIST3D and AdrenalMNIST3D, we used a small MLP trained directly on the encoded shape coordinates (see Section~\ref{sec:result-feret}).
This MLP consists of three layers with 256 channels, yielding 132,353 trainable parameters.
For all MedMNIST3D datasets, we used the official data splits.

\subsubsection{EchoNet.}
Although videos are often treated as 3D volumes \cite{ouyang_video-based_2020}, they exhibit a spatio-temporal structure characterized by \textit{dimensionally decoupled anisotropy} - that is, the spatial and temporal dimensions do not share a common metric space.
As a result, distances or relationships along the temporal axis are fundamentally different from those in the spatial domain, necessitating PEs that account for this anisotropy.

The EchoNet Dynamic dataset\footnote{\url{https://echonet.github.io/dynamic/}} contains apical-4-chamber echocardiography videos of different lengths, with a training set of 7,460 videos, a validation set of 1,288, and a test set of 1,276 \cite{ouyang_video-based_2020}.
The dataset consists of 48\% female and 52\% male patients with an average age of 68 years.
The videos were downsampled by cubic interpolation into standardized \(112^2\) pixel videos.
During training, \(16\) consecutive frames were randomly sampled.
The patch size was chosen to be \(2\times8^2\) (frames, height, width), resulting in \(8\times14^2\) patches for every video.
The goal in this dataset is to predict the EF, which is expressed as a percentage and is the ratio of left ESV and left EDV: \(EF(\%)=\frac{EDV-ESV}{EDV} \times 100\).

The performance was evaluated using the \(R^2\) score metric as in the original publication~\cite{ouyang_video-based_2020}.
The ViTs trained on this dataset were trained with 3 layers, 6 heads and a MLP dimension of 768 resulting in 11.1 million trainable parameters.

\subsection{Training, Optimization and Evaluation}
All models trained on ChestX and EchoNet were trained for 150 epochs on the training dataset while exploring different hyperparameter configurations.
For SPE, we evaluated temperatures $t \in \{10^3, 10^4\}$.
For IFPE, we tested scale values $s \in \{0.5, 1.0, 5.0\}$. 
For AFPE, we employed random sampling to determine the optimal $s_i$ values.
Specifically, for each dataset, we performed 50 training runs with randomly sampled $s_i$ values, each trained for 75 epochs.
The optimal hyperparameters were selected based on validation set performance.

For the experiments on OrganMNIST3D, no hyperparameter optimization was performed.
The AFPE scale parameters were set according to the anisotropy of the data.
Specifically, for each spatial dimension $i$, the scale was defined as $s_i = 0.5 \times \text{anisotropy}_i$.
IFPE and SPE parameters have been set to the commonly used $t = 10^4$ and $s=1.0$ respectively.

In all experiments, the final performance metrics were reported on the test sets.
To ensure robust evaluation, we performed 50 bootstrap runs and reported the mean and standard deviation of the resulting metrics.
To assess statistical significance, we performed a t-test comparing the means of the bootstrapped performance.
The null hypothesis assumed that repeated samples had identical means.
We rejected the null hypothesis if the resulting $p$-value was less than 0.05.

Performance metrics were chosen to reflect standards in the field \cite{maier-hein_metrics_2024}, with implementations taken from TorchMetrics (version 1.5.2)~\cite{nicki_skafte_detlefsen_torchmetrics_2024}.
All models are implemented using Python (version 3.10.15), MONAI \cite{cardoso_monai_2022} (version 1.4.0) and PyTorch (version 2.4.0) and trained on a NVIDIA A100 40G GPU.
The code is made publicly available as referenced in the abstract.

\section{Results and Discussion}
\subsection{AFPE Encodes Anatomical Shape Priors for Classification}
\begin{figure}[h]
    \centering
    \includegraphics[width=\linewidth]{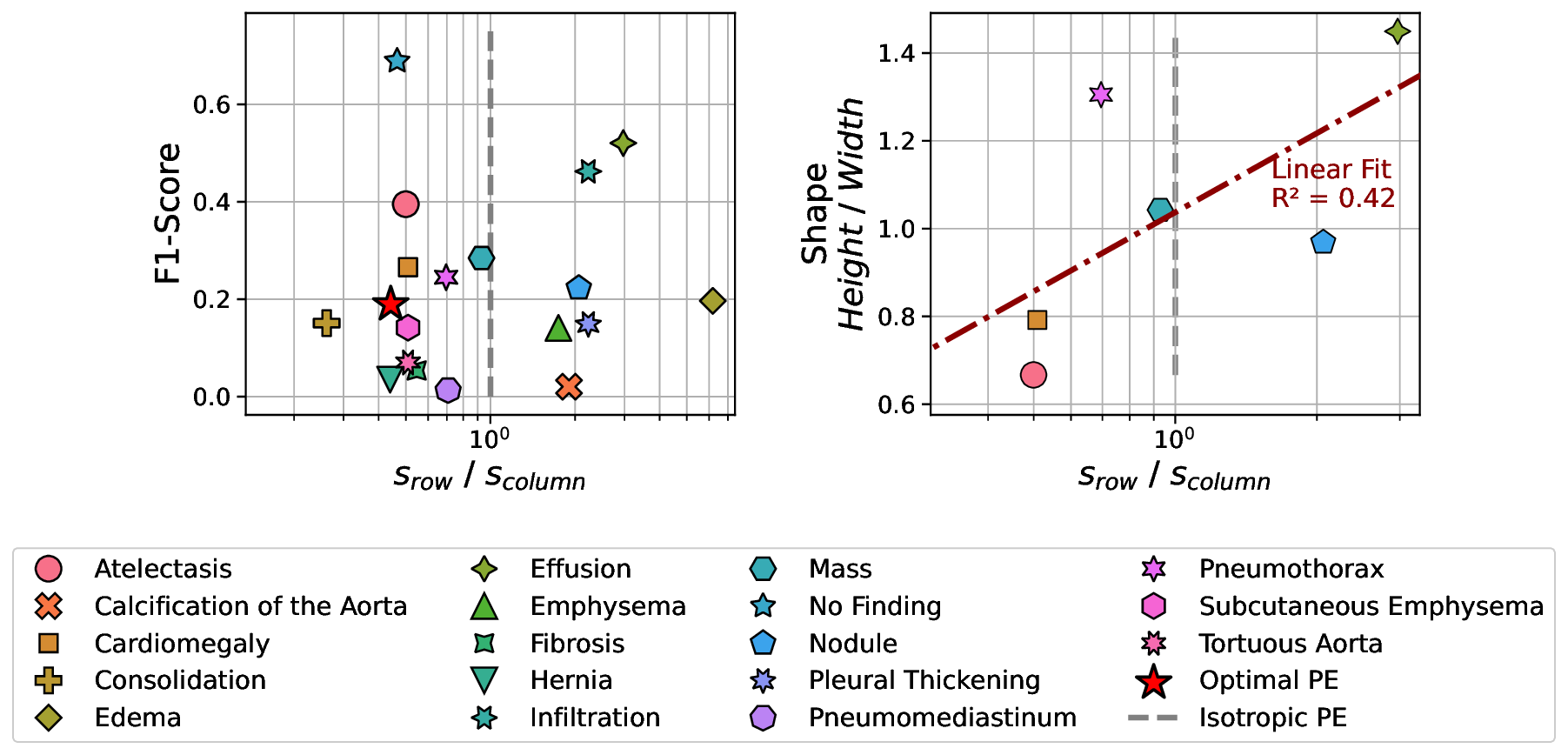}
    \caption{Optimal AFPE depends on the target class and enables incorporation of anatomical shape priors. The left subplot shows the isotropy and performance of the best-performing model on the ChestX validation data for each class. The right subplot shows the relationship between isotropy and bounding box shape for the ChestX subset that includes bounding box annotations. A linear regression line (red) is fitted, and the corresponding $R^2$ is reported.
    }
    \label{fig:optimal-afpe}
\end{figure}

In the ChestX dataset two different AFPE scale hyperparameters $s_{row}$ and $s_{column}$ were tuned. 
The effect of changing these hyperparameters on the resulting AFPE can be seen in Figure~\ref{fig:iso-pe-comparison-2d}.
The final AFPE model for ChestX has $s_{row}=0.497$ and $s_{column}=1.125$.
As shown in Figure~\ref{fig:optimal-afpe} (left), the optimal ratio of $s_{row}$ and $s_{column}$ depends on the label being classified.
It can be observed that when $s_{row} < s_{column}$ AFPE performs well on diseases associated with airway collapse (Atelectasis, Pneumothorax), cardiac associated diseases (Cardiomegaly, Tortuous Aorta), and air escape disorders (Subcutaneous Emphysema, Pneumomediastinum).
On the other hand, when $s_{row} > s_{column}$, conditions like Emphysema, Pleural Thickening, Infiltration, effusion, and Nodule suggest a focus on lung tissue damage, pleural changes, and inflammatory or fibrotic responses.
It does align with our expectation that Mass is optimally predicted when $s_{row} = s_{column}$, likely because of its round shape.
Overall, tuning $s_{row}$ and $s_{column}$ improved the detection of specific diseases by aligning shape priors with anatomical patterns.
For example, in diseases with predominantly vertical patterns, such as cardiomegaly, the best performing model has a reduced horizontal similarity ($s_{row} < s_{column}$).
This aligns with the diagnostic approach to cardiomegaly, where an enlarged cardiac silhouette and a cardiothoracic ratio exceeding 50\% indicate the condition~\cite{mensah_establishing_2015}.

Even though often the case, the diagnostic axis for a particular disease does not always align with the longest anatomical axis (Figure~\ref{fig:optimal-afpe} right).
In pneumothorax, for example, diagnostic bounding boxes typically span a large portion of one lung.
Despite the lung’s primary vertical orientation, key diagnostic features - such as the visceral pleural line and absence of peripheral vascular markings - often present along a horizontal axis \cite{sakai_thoracic_2020}. Regardless of whether the anatomical shape prior concerns the diagnostic axis or the overall geometry, AFPE enables alignment to the required shape.

\subsection{AFPE Outperforms Canonical Positional Encodings under Anisotropy}
Table \ref{tab:main-table} compares the performance of the proposed AFPE with canonical PEs across three different datasets.
In all settings with anisotropy greater than one, AFPE consistently outperforms the other PEs.
In the OrganMNIST3D dataset, a substantial drop in performance is observed upon the introduction of anisotropy.
However, AFPE demonstrates a stronger ability to model anisotropic structures, mitigating the degradation in performance more effectively than the alternatives.

The most pronounced performance improvement of AFPE over other methods is observed on the EchoNet dataset.
These videos exhibit high anisotropy, characterized by spatial and temporal dimensions that do not share a common metric space, unlike 3D volumetric images.
A more detailed analysis of these findings and their implications for the EchoNet dataset is presented in the following section.

\begin{table}[ht]
    \centering
    \caption{
    Performance comparison of AFPE and canonical PEs on isotropic and anisotropic datasets. Mean $\pm$ standard deviation over 50 bootstrap runs on test data are reported. Best results with statistically significant ($p<0.05$) improvement are \textbf{bolded}, otherwise the top two results are \underline{underlined}.
    "None" indicates no PE, "Learnable" indicates learned PE. Input and patch sizes reflect the spatial dimensions used for each dataset.
    }
    \label{tab:main-table}
\begin{tabular}{rcccccccc}
\toprule
                           & ChestX                        &  & \multicolumn{4}{c}{OrganMNIST3D}                                                                                              &  & EchoNet                       \\ \cline{2-2} \cline{4-7} \cline{9-9} 
Anisotropy                  & 1                             &  & 1                             & 4                             & 6                           & 8                             &  & 7                             \\
Input size                 & $224^2$                       &  & $64^3$                        & $16\times64^2$                & $11\times64^2$                & $8\times64^2$                 &  & $16\times112^2$               \\
Patch size                 & $16^2$                        &  & $4^3$                         & $1\times4^2$                  & $1\times4^2$                  & $1\times4^2$                  &  & $2\times8^2$                  \\
Metric                     & AUPRC $\uparrow$              &  & \multicolumn{4}{c}{AUROC $\uparrow$}                                                                                          &  & $R^2$ $\uparrow$              \\ \cline{2-2} \cline{4-7} \cline{9-9} 
\multicolumn{1}{l}{Method} &                               &  &                               &                               &                               &                               &  &                               \\
None                       & \pmnum{0.146}{0.002}          &  & \pmnum{0.992}{0.001}          & \pmnum{0.967}{0.003}          & \pmnum{0.964}{0.003}          & \pmnum{0.959}{0.003}          &  & \pmnum{0.283}{0.034}          \\
Learnable                  & \pmnum{0.156}{0.002}          &  & \pmnum{0.983}{0.002}          & \pmnum{0.953}{0.004}          & \pmnum{0.951}{0.003}           & \pmnum{0.944}{0.004}          &  & \pmnum{0.334}{0.029}          \\
SPE                        & \underline{\pmnum{0.186}{0.003}} &  & \textbf{\pmnum{0.995}{0.001}} & \underline{\pmnum{0.980}{0.003}}          & \underline{\pmnum{0.975}{0.003}}         & \underline{\pmnum{0.969}{0.003}}          &  & \pmnum{0.527}{0.028}          \\
IFPE                        & \pmnum{0.184}{0.003}          &  & \underline{\pmnum{0.994}{0.001}}          & \pmnum{0.979}{0.003}          & \underline{\pmnum{0.975}{0.003}}          & \pmnum{0.958}{0.004}          &  & \underline{\pmnum{0.547}{0.027}}          \\
LFPE                       & \pmnum{0.179}{0.003}          &  & \pmnum{0.993}{0.001}          & \pmnum{0.969}{0.003}          & \pmnum{0.972}{0.003}          & \underline{\pmnum{0.969}{0.003}}          &  & \pmnum{0.522}{0.029}          \\
\textbf{AFPE}              & \underline{\pmnum{0.185}{0.004}}          &  & \underline{\pmnum{0.994}{0.001}}          & \textbf{\pmnum{0.984}{0.002}} & \textbf{\pmnum{0.983}{0.002}} & \textbf{\pmnum{0.972}{0.003}} &  & \textbf{\pmnum{0.621}{0.024}} \\ \bottomrule
\end{tabular}

\end{table}

\subsection{AFPE Separately Encodes Spatial and Temporal Information}

\begin{figure}[ht]
    \centering
    \includegraphics[width=\linewidth]{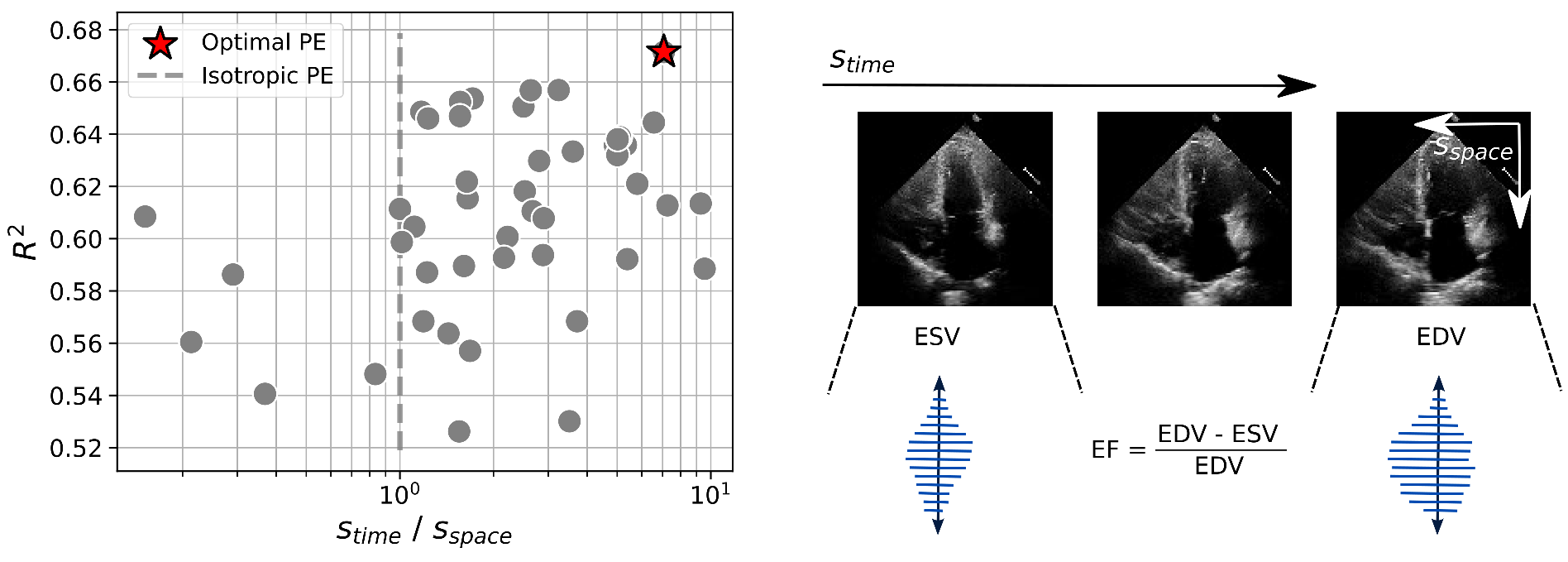}
    \caption{Left: Hyperparameter tuning results for AFPE on the EchoNet validation set. Right: Schematic illustration of the EchoNet task. In this setting, spatial and temporal dimensions serve distinct roles in estimating the ejection fraction (EF). Accurate EF prediction requires identifying the correct frames corresponding to end-systolic volume (ESV) and end-diastolic volume (EDV), followed by estimating their respective volumes.}
    \label{fig:echonet}
\end{figure}

In the EchoNet dataset, we tuned the AFPE hyperparameters \( s_{\text{time}} \) and \( s_{\text{space}} \), where \( s_{\text{space}} \) was constrained to be identical across both spatial dimensions.
As illustrated in Figure~\ref{fig:echonet} (right), \( s_{\text{time}} \) controls similarity along the temporal dimension, while \( s_{\text{space}} \) governs similarity in space.
The final model uses \( s_{\text{time}} = 0.785 \) and \( s_{\text{space}} = 0.111 \), reflecting greater independence (i.e., lower similarity) across temporal frames.
Figure~\ref{fig:echonet} (left) shows the performance of all tested scale parameters on the validation set.

In Figure~\ref{fig:echonet} (right), the complexity of the EchoNet task is shown schematically.
It arises from the need to first identify the relevant frames corresponding to ESV and EDV, and then estimate their volumes in order to finally compute the EF.

As shown in Table~\ref{tab:main-table}, IFPE is the second-best performing PE in the EchoNet task, suggesting that Fourier features are particularly well-suited for modeling temporal dynamics in this context.
However, the substantial performance gain achieved by AFPE over IFPE highlights the importance of independently optimizing spatial and temporal similarity through the parameters \( s_{\text{space}} \) and \( s_{\text{time}} \).
The significance of effective spatio-temporal encoding is further emphasized by the notably lower performance of the baseline model without any PE (\textit{None}), which underperforms all PE-based models.
Note that this baseline does perform well on OrganMNIST3D, where texture likely plays a larger role than spatial composition.

Two hypotheses may explain the significant performance improvement observed in EchoNet.
First, the dimensionally decoupled anisotropy in video-based AFPE allows it to better exploit the structure of the data.
Second, Fourier features have been shown to effectively represent complex shapes~\cite{tancik_fourier_2020}, which likely also applies to EDV and ESV.
To further evaluate the second hypothesis, the following section investigates the ability of SPE, IFPE, and AFPE to learn high-level shape descriptors from spatial coordinates alone.

\subsection{AFPE \& IFPE Encode High Level Shape Descriptors}
\label{sec:result-feret}
As illustrated in Figure~\ref{fig:echonet} and described in \cite{ouyang_video-based_2020}, ESV and EDV can be approximated using Feret's maximum diameter (indicated by the blue arrow) along with several perpendicular measurements.
To evaluate the hypothesis that Fourier-based encodings are better suited than SPE for capturing shape descriptors, we conducted experiments on the AdrenalMNIST3D and VesselMNIST3D datasets.

In both datasets, the shapes were encoded by directly applying either SPE (Eq.~\ref{eq:spe}), IFPE (Eq.~\ref{eq:fpe}), or AFPE (Eq.~\ref{eq:afpe}) on the coordinates of the shapes.
Given these encoded positions, the task was to predict the minimum and maximum Feret diameter under varying anisotropies using a simple 3-layer MLP.
The results, summarized in Table~\ref{tab:fermet}, demonstrate that IFPE and AFPE consistently outperform SPE.

Previous work has encoded 3D shapes in function space using occupancy networks~\cite{mescheder_occupancy_2019}.
Tancik et al.~\cite{tancik_fourier_2020} demonstrated that Fourier features significantly enhance the learning of high-frequency functions, such as those required to represent complex shapes.
Our results are consistent with these findings, showing that Fourier-based encodings enable neural networks to learn high-frequency shape descriptors that are relevant for medical annotation \cite{ouyang_video-based_2020}.

Given that IFPE and AFPE perform comparably across all evaluated settings in Table~\ref{tab:fermet}, we conclude that their respective capabilities in encoding anatomical shape are not significantly different.
Therefore, the substantial performance gap observed between the two methods in the EchoNet benchmark (Table~\ref{tab:main-table}) is likely attributable to the dimensionally decoupled anisotropy inherent to video data, rather than differences in shape encoding.

\begin{table}[ht]
\centering
\caption{Performance comparison of PE methods in predicting minimum and maximum Feret’s Diameter (FD) across different anisotropy settings. Results are reported as mean $\pm$ standard deviation over 100 bootstrap runs on the test data. The best-performing results per setting are \textbf{bolded}.}
\label{tab:fermet}
\begin{tabular}{rcccccc}
\toprule
     & \multicolumn{1}{l}{} & \multicolumn{2}{c}{AdrenalMNIST3D $R^2 \uparrow$}              &  & \multicolumn{2}{c}{VesselMNIST3D $R^2\uparrow$}               \\ 
     & \multicolumn{1}{l}{} & Min. FD                        & Max. FD                        &  & Min. FD                        & Max. FD                        \\ \cline{3-4} \cline{6-7}
\multicolumn{1}{l}{Method}    & Anisotropy           &                               &                               &  &                               &                               \\ \hline
SPE  &                      & \pmnum{0.071}{0.067}          & \pmnum{0.750}{0.083}          &  & \pmnum{0.076}{0.026}          & \pmnum{0.061}{0.036}          \\
IFPE & 3                    & \textbf{\pmnum{0.725}{0.032}} & \pmnum{0.921}{0.010}          &  & \textbf{\pmnum{0.635}{0.025}} & \pmnum{0.913}{0.014}          \\
AFPE &                      & \pmnum{0.683}{0.031}          & \textbf{\pmnum{0.922}{0.013}} &  & \pmnum{0.624}{0.0.28}         & \textbf{\pmnum{0.919}{0.009}} \\ \hline
SPE  &                      & \pmnum{0.132}{0.059}          & \pmnum{0.711}{0.055}          &  & \pmnum{0.060}{0.024}          & \pmnum{0.101}{0.032}           \\
IFPE & 5                    & \pmnum{0.681}{0.045}          & \textbf{\pmnum{0.885}{0.014}} &  & \pmnum{0.527}{0.036}          & \textbf{\pmnum{0.902}{0.015}} \\
AFPE &                      & \textbf{\pmnum{0.717}{0.032}} & \pmnum{0.880}{0.019}          &  & \textbf{\pmnum{0.587}{0.035}} & \pmnum{0.869}{0.014}          \\\bottomrule        
\end{tabular}
\end{table}

\section{Conclusion and Outlook}
This work introduces AFPE, a novel approach for encoding spatial and temporal coordinates in medical imaging tasks with geometric complexity.
By generalizing IFPE to allow for directional control over positional similarity, AFPE enhances neural networks' ability to model anisotropic structures - particularly relevant in 3D and time-resolved medical data.

Our findings demonstrate that AFPE consistently enhances performance across diverse tasks and imaging modalities, particularly in scenarios where anatomical shape, orientation, and motion are critical.
The benefits of AFPE is especially pronounced in dimensionally decoupled and anisotropic settings, such as spatio-temporal data.
These results highlight the potential of geometry-informed priors to significantly improve both classification and regression performance in medical imaging applications.

Future directions include applying AFPE to segmentation models such as SAM~\cite{kirillov_segment_2023} and object detection frameworks like DETR~\cite{carion_end--end_2020}, as well as integrating scale parameters into end-to-end learning pipelines to improve efficiency and adaptability.
Ultimately, enhancing inductive shape bias through PE may reduce data requirements and improve robustness in clinical deployments.
\newline

\noindent \textbf{Acknowledgements.} This work was supported by the German Ministry for Research, Technology and Space (BMFTR, junior research group “Medical Omics”, 01ZZ2001).

\bibliographystyle{splncs04}
\bibliography{references}





\end{document}